\documentclass{article}

\PassOptionsToPackage{numbers}{natbib}
\usepackage[main, final]{neurips_2026}

\usepackage[utf8]{inputenc}
\usepackage[T1]{fontenc}
\usepackage{hyperref}
\usepackage{url}
\usepackage{booktabs}
\usepackage{amsfonts}
\usepackage{nicefrac}
\usepackage{microtype}
\usepackage{xcolor}

\usepackage{graphicx}
\usepackage{amsmath}

\title{AdaTree: Serving-Aware Adaptive Tree Construction for Speculative Decoding}

\author{%
  Soohyun Cha\thanks{The author is now at Samsung Research (soohyun.cha@samsung.com).}\\
  Seoul National University \\
  \texttt{soohyun.cha@snu.ac.kr} \\
  \And
  Jungi Lee \\
  Seoul National University \\
  \texttt{jungi.lee@snu.ac.kr} \\
  \And
  Jaewoong Sim \\
  Seoul National University \\
  \texttt{jaewoong@snu.ac.kr} \\
}

\usepackage{pdfprivacy}
\usepackage{xspace}

\newcommand{\putsec}[2]{\section{#2}\label{sec:#1}}
\newcommand{\putssec}[2]{\subsection{#2}\label{ssec:#1}}
\newcommand{\putsssec}[2]{\subsubsection{#2}\label{sssec:#1}}

\newcommand{\ssecref}[1]{Section~\ref{ssec:#1}}
\newcommand{\sssecref}[1]{Section~\ref{sssec:#1}}

\newcommand{\myparagraph}[1]{\noindent\textbf{#1}}

\newcommand{\figref}[1]{Figure~\ref{fig:#1}}
\newcommand{\tabref}[1]{Table~\ref{tab:#1}}
\newcommand{\eqnref}[1]{Equation~\ref{eqn:#1}}

\newcommand{\name}{{AdaTree\xspace}}
\newcommand{\mm}[2]{\ensuremath{{#1}_\mathrm{#2}}}

\begin{document}

\maketitle

\begin{abstract}

Speculative decoding accelerates generative inference of large language models
(LLMs) by using a small draft model to propose multiple candidate tokens, which
are then verified in parallel with the target model in a single decoding
iteration.
While the state-of-the-art method of tree-based speculative decoding helps
improve generation throughput over non-speculative inference, deploying it in
LLM serving systems often yields suboptimal performance due to dynamically
changing serving conditions.
In particular, our analysis shows that the optimal tree configuration---the one
that maximizes performance---varies with two key serving conditions: request
rate and per-request characteristics.
Based on the analysis, we present \name{}, a plug-in component for LLM serving
systems that dynamically adapts tree configurations to varying serving
conditions.
\name{} predicts model execution time and acceptance length across different
tree configurations and selects the one that maximizes speculative decoding
efficiency.
To capture the non-linear relationship between tree configuration and model
execution time, \name{} employs decision trees as its core modeling primitive.
Our evaluation shows that \name{} consistently outperforms both chain-based and
static tree-based speculative decoding across diverse serving conditions.

\end{abstract}

\putsec{intro}{Introduction}

Speculative decoding (SD) is a popular technique that alleviates the sequential
bottleneck of autoregressive generation in large language models (LLMs) through
a two-stage decoding approach~\cite{che:bor23,lev:kal23}.
In the first stage, a lightweight draft model rapidly proposes multiple
candidate tokens.
In the second stage, the target model verifies these candidates \emph{in
parallel} and accepts the longest prefix consistent with its own output
distribution.
This allows multiple tokens to be generated within a single forward pass of the
target model, thereby accelerating LLM inference.
For drafting, state-of-the-art methods organize candidate tokens as a
tree~\cite{cai:li24,che:may24,li:wei24-eagle2,mia:oli24} rather than as a
chain~\cite{che:bor23} to increase the number of accepted tokens per iteration
(i.e., the acceptance length).

However, prior work typically relies on fixed tree configurations, limiting its
effectiveness in practical serving environments where the request rate and
per-request characteristics fluctuate over time.
For example, when batch sizes are small and compute resources are underutilized,
drafting larger token trees would be advantageous as they can potentially
increase the acceptance length with a negligible increase in verification
overhead.
Conversely, when batch sizes are large and compute utilization is already high,
larger trees would substantially increase verification time; in such cases,
smaller trees may provide better overall performance despite accepting fewer
tokens per iteration.
Similarly, for requests that are easy for the draft model to predict, a deeper
tree would increase the acceptance length, as longer candidate sequences are
likely to be accepted more often; for harder requests, however, it would merely
add drafting overhead.

In this paper, we first analyze the relationship between the optimal token tree
structure and two key serving conditions: request rate and request
characteristics (e.g., token predictability).
We observe that dynamically adjusting the tree configuration in response to
fluctuations in these conditions is crucial for maximizing the performance
benefits of speculative decoding.

Motivated by this observation, we propose \name{}, a plug-in component for LLM
serving systems that adapts the token tree structure to changing serving
conditions.
\name{} predicts model execution time and acceptance length for different tree
configurations and selects the one that maximizes speculative decoding
efficiency.
Specifically, it employs decision trees to capture the non-linear relationship
between the tree configuration and model execution time.
We integrate \name{} into the official codebase of a state-of-the-art tree-based
SD method~\cite{li:wei25}, which we extend with continuous
batching~\cite{yu:jeo22}, and evaluate it under diverse serving conditions.
Our results show that \name{} consistently outperforms baseline approaches that
use static configurations for speculative decoding.

Our contributions are as follows:

\begin{itemize}
  \item We analyze how the optimal tree structure varies with request rate and
    per-request characteristics, and show that adapting it to these serving
    conditions is essential for maximizing SD performance, particularly in
    long-context scenarios.
  \item We propose \name{}, a plug-in software component for LLM serving systems
    that automatically adjusts the size and depth of the token tree.
    It uses XGBoost~\cite{che:gue16} to predict model execution time, which
    exhibits non-linear characteristics.
  \item We demonstrate that an adaptive token tree accelerates generative
    inference in fluctuating serving environments compared to SD with a static
    tree structure.
\end{itemize}

\putsec{background}{Background}

\putssec{back-sd}{Speculative Decoding}

Speculative decoding (SD)~\cite{che:bor23,lev:kal23} is a lossless LLM
acceleration technique that repeatedly drafts multiple tokens with a lightweight
draft model and verifies them in parallel with the base LLM, referred to as the
target model.
We denote the draft and target models as $M_{D}$ and $M_{T}$, respectively.

Early SD work drafts tokens sequentially, forming a token
chain~\cite{che:bor23}, and subsequent work has explored diverse drafting
strategies.
BiLD~\cite{kim:man23} stops drafting when the maximum prediction probability of
the draft model falls below a threshold.
Kangaroo~\cite{liu:tan24} adopts a similar early-stopping policy but uses a
shallow sub-network of the target model as the draft model.
MagicDec~\cite{sad:che25} uses a compressed KV cache for drafting to keep its
cost low in long-context, large-batch scenarios.

\putssec{tree-sd}{Token Tree-based SD}

Instead of organizing draft tokens as a chain, subsequent work also organizes
them in a tree structure to increase the number of accepted tokens per
iteration.
SpecInfer~\cite{mia:oli24} introduces tree-mask attention with an ensemble of
draft models.
Medusa~\cite{cai:li24} attaches extra decoding heads to the target model to
predict multiple future tokens in parallel.
Sequoia~\cite{che:may24} uses dynamic programming to find an optimal tree
structure offline.
EAGLE~\cite{li:wei24-eagle1} advances tree-based SD with a new draft model
structure and training method.

Some recent work constructs draft trees dynamically rather than relying on a
predefined tree topology.
GLIDE with a CAPE~\cite{du:jia24} adjusts the number of draft tokens based on
the draft token probability while expanding along only a single fixed branch.
EAGLE-2~\cite{li:wei24-eagle2} selectively expands branches at runtime,
achieving state-of-the-art performance~\cite{xia:yan24}.
EAGLE-3~\cite{li:wei25} extends EAGLE-2 with a new training method.

\begin{figure} [t]
    \centering
    \includegraphics[width=\linewidth]{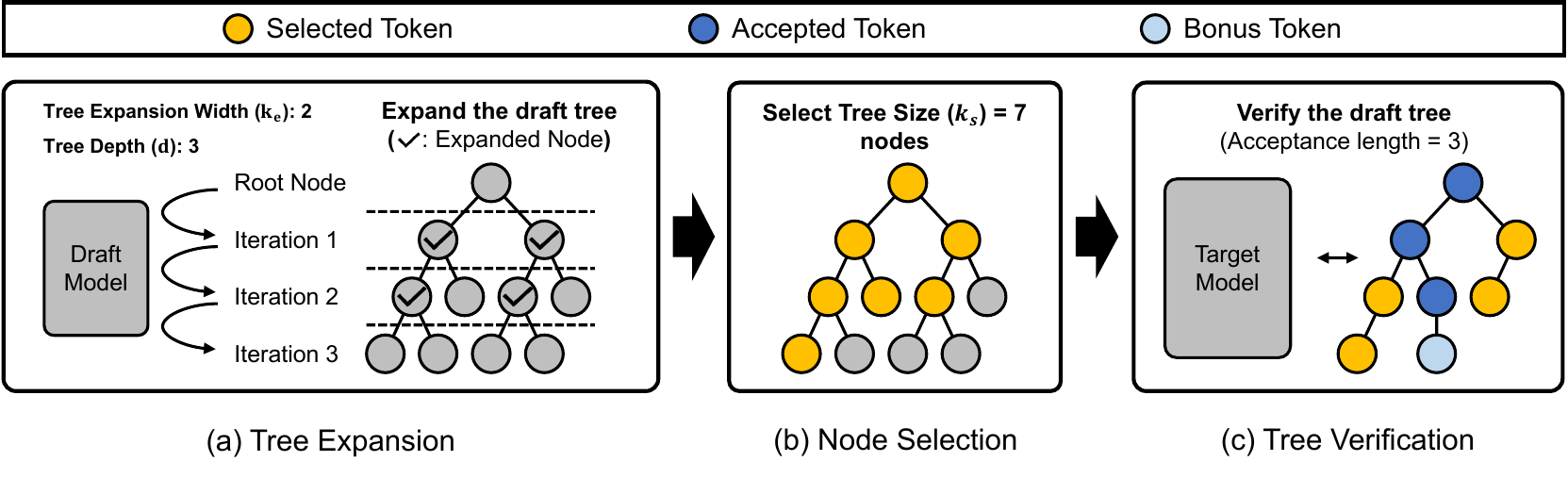}
    \vspace{-0.25in}
    \caption{Draft tree construction and verification in EAGLE-2.
      (a) Tree expansion through successive forward passes of the draft model.
      (b) Node selection to form the final draft tree.
      (c) Tree verification by the target model.}
    \vspace{-0.10in}
    \label{fig:back-eagle2}
\end{figure}

\figref{back-eagle2} illustrates the tree construction process of EAGLE-2.
As shown in \figref{back-eagle2}(a), it first selectively expands branches for
the top-\mm{k}{e} (e.g., 2 in the figure) nodes with the highest probabilities
from the draft model.
Once the tree reaches the maximum depth $d$ (e.g., 3 in the figure), it selects
the top-\mm{k}{s} nodes (e.g., 7 in the figure) with the highest probabilities,
as shown in \figref{back-eagle2}(b).
Finally, as shown in \figref{back-eagle2}(c), the target model verifies all
selected nodes in parallel, accepts the tokens with correct outputs, and
generates one bonus token.
The \textit{acceptance length} is defined as the number of accepted tokens,
including the bonus token but excluding the root token, which the target model
has already generated.
We further define \textbf{\textit{tree expansion width}} as the number of nodes
expanded at each level (i.e., \mm{k}{e}), \textbf{\textit{tree size}} as the
number of tokens in the tree after node selection (i.e., \mm{k}{s}), and
\textbf{\textit{tree depth}} as the maximum distance from the root node to any
leaf node (i.e., $d$).

\putssec{serving-systems}{LLM Serving Systems}

Modern LLM serving systems employ \textit{continuous batching}~\cite{yu:jeo22},
a fine-grained batching technique that adds new requests and removes completed
ones at every iteration.
Under continuous batching, both the batch size and the request characteristics
can fluctuate over time.
Some prior work~\cite{su:gia23,qia:gon24,wu:zho25,li:zha26} adaptively adjusts
the number of draft tokens based on the batch size, but these methods are
designed specifically for chain-based drafting and do not directly extend to
tree-based SD, which requires jointly adapting tree size and depth.

\putsec{motivation}{Characterization of Tree-based Speculative Decoding}

For a given tree construction algorithm, two key parameters influence the
performance of SD: \textit{\textbf{tree size}} and \textit{\textbf{depth}}.
Both affect the acceptance length, but they incur different costs: tree size
mainly affects the target model's verification time, whereas tree depth mainly
affects the draft model's execution time.
In this section, we analyze the performance of tree-based SD by varying these
two parameters.
We show that their optimal values, where performance is maximized, vary with two
serving conditions: \textit{request rate} and \textit{request characteristics}.
Furthermore, the performance gap between suboptimal and optimal tree
configurations widens in long-context scenarios, further motivating the need for
adaptive tree construction.
Implementation details are described in~\ssecref{setting}.

\putssec{motiv-size}{Optimal Tree Size Varies with Request Rate}

As the request rate increases, more requests are aggregated into each batch.
The target model verifies the tree nodes of all requests in the batch in
parallel, while the draft model sequentially expands each tree one level at a
time.
Thus, the optimal tree size varies with the request rate, since a higher request
rate results in more trees to be verified in parallel.

\begin{figure}[t]
    \centering
    \includegraphics[width=\linewidth]{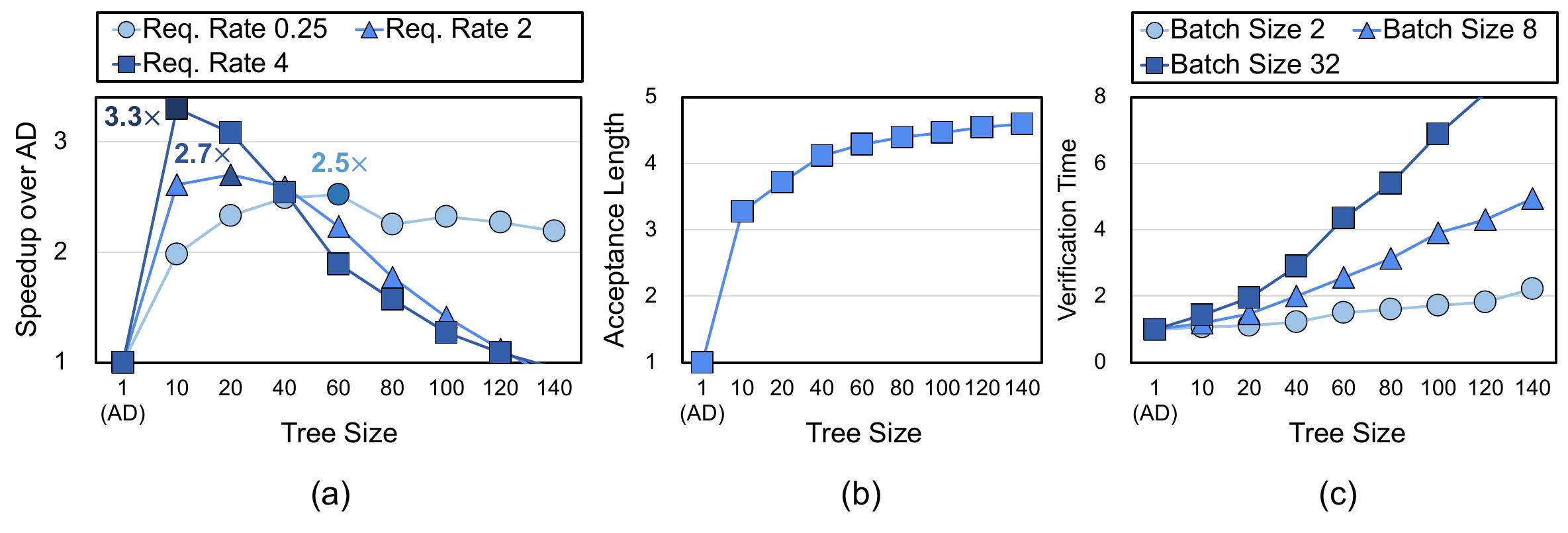}
    \vspace{-0.25in}
    \caption{(a) Speedup of tree-based SD over AD under varying tree sizes for
      three different request rates.
      (b) Acceptance length of tree-based SD across tree sizes.
      (c) Verification time normalized to that of AD for three batch sizes,
      representing the average batch size at each request rate.
    }
    \vspace{-0.15in}
    \label{fig:char-system-load}
\end{figure}

\figref{char-system-load} presents the speedup of tree-based SD over
autoregressive decoding (AD) across varying tree sizes and request rates, which
we analyze in terms of acceptance length and verification time.
For the experiments, we use Llama-3.1-8B-Instruct~\cite{gra:dub24} as the target
model with its EAGLE draft model, the GSM8K dataset~\cite{cob:kos21}, and an
NVIDIA RTX Pro 6000 Blackwell GPU.

\myparagraph{Benefit of Increasing Tree Size.}
The speedup initially increases with tree size.
This is because a larger tree covers more candidate continuations and thus
increases the acceptance length, as shown in
Figures~\ref{fig:char-system-load}(a) and (b).

\myparagraph{Computational Overhead of Increasing Tree Size.}
Beyond a certain size, however, the speedup decreases because verification time
grows linearly with tree size once verification shifts from memory
bandwidth-bound to compute-bound, as shown in
Figures~\ref{fig:char-system-load}(a) and (c).
Consequently, the computational overhead eventually outweighs the benefit of
improved acceptance length.

\myparagraph{Variation with Different Request Rates.}
At a request rate of 0.25~requests/s (average batch size of 2), the speedup
continues to increase up to a tree size of 60.
This is because verification remains memory bandwidth-bound up to this size, so
its latency stays comparable to that of AD, as shown
in~\figref{char-system-load}(c).
Conversely, at a higher request rate (e.g., 4~requests/s), larger batches (e.g.,
average batch size of 32) cause the verification to saturate into a
compute-bound regime more rapidly as the tree size grows, since the target model
needs to verify more trees in parallel.
Meanwhile, the acceptance length remains unchanged across batch sizes, so the
benefit of SD does not increase with larger batches.
As a result, as shown in \figref{char-system-load}(a), the optimal tree size
decreases as the batch size grows with higher request rates.
This implies that the tree size should be dynamically adjusted based on the
request rate.

\putssec{motiv-depth}{Optimal Tree Depth Varies with Request Characteristics}

\begin{figure}[b]
    \centering
    \includegraphics[width=\linewidth]{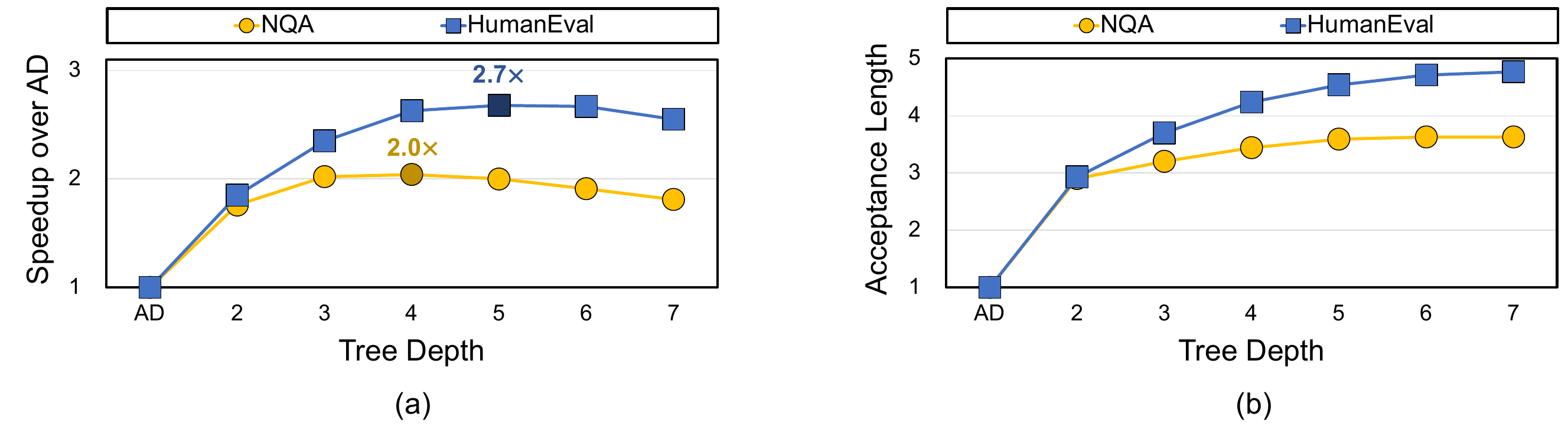}
    \vspace{-0.15in}
    \caption{
      (a) Speedup of tree-based SD over AD across tree depths, evaluated on two
      different datasets, NQA and HumanEval.
      (b) Acceptance length of tree-based SD across varying depths.
    }
    \vspace{-0.10in}
    \label{fig:char-req-semantics}
\end{figure}

The optimal tree depth varies with request characteristics because the
acceptance length, the primary benefit of using tree-based SD, depends on how
predictable each request is for the draft model.
\figref{char-req-semantics} presents the speedup of tree-based SD over AD across
varying tree depths on two datasets with distinct categories:
NQA~\cite{kwi:pal19} for question answering and HumanEval~\cite{che:two21} for
code completion.

\myparagraph{Benefit of Increasing Tree Depth.}
The speedup increases with tree depth up to a certain point.
This is because a deeper tree increases the acceptance length by allowing longer
draft sequences to be accepted.

\myparagraph{Computational Overhead of Increasing Tree Depth.}
The acceptance length, however, eventually saturates because the draft model
fails to predict certain tokens due to its limited capacity.
Consequently, the speedup decreases at greater depth, making the additional runs
of the draft model wasteful.

\myparagraph{Variation with Different Request Characteristics.}
As shown in \figref{char-req-semantics}(a), the speedup continues to increase up
to a depth of 5 for HumanEval, whereas it peaks at 4 for NQA.
This is because the acceptance length for HumanEval grows more sharply at depth
4 than for NQA, as shown in \figref{char-req-semantics}(b), making additional
draft model runs less beneficial for NQA.
The longer acceptance length in HumanEval is likely due to its nature as a code
completion task, which contains more structured and recurring subsequences that
are easier for the draft model to predict, whereas NQA, a question answering
task, has a more diverse context, resulting in a lower acceptance
length~\cite{che:bor23}.
This implies that the tree depth should be dynamically adjusted based on
per-request characteristics.

\putssec{motiv-long-context}{Growing Necessity of Optimal Tree in Long-Context Serving Scenarios}

\begin{figure}[t]
    \centering
    \includegraphics[width=\linewidth]{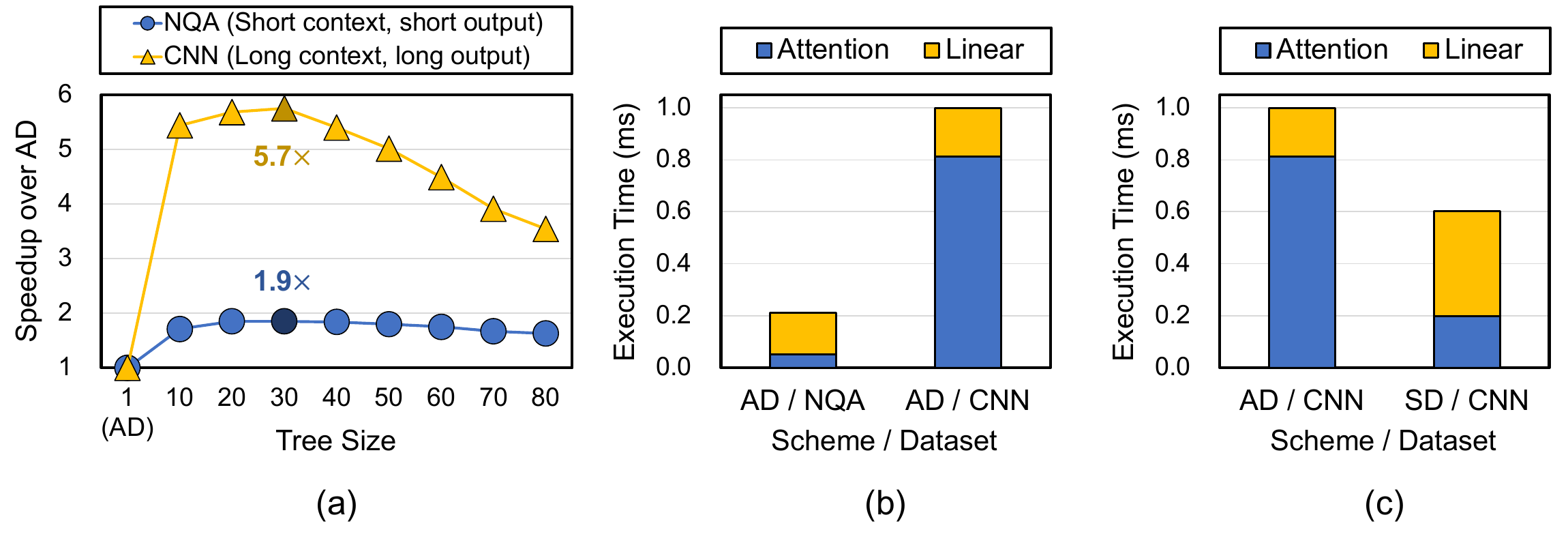}
    \vspace{-0.25in}
    \caption{
      (a) Speedup of tree-based SD over AD under varying tree sizes, evaluated
      on the two datasets (NQA and CNN) under 1.0 requests/sec.
      (b) Execution time breakdown of AD for two datasets.
      (c) Execution time breakdown of CNN for AD and SD with an optimal tree
      size of 30.
    }
    \vspace{-0.10in}
    \label{fig:char-long-context}
\end{figure}

Recent LLM workloads increasingly involve reasoning
models~\cite{jae:kal24,yan:li25}, which produce long chain-of-thought outputs,
and agentic workflows~\cite{wan:ma24}, which accumulate long contexts over
multi-turn interactions.
In such long-context scenarios, constructing an optimal tree becomes more
critical, as a growing key-value (KV) cache amplifies the memory bandwidth
bottleneck during token generation.

To illustrate this, \figref{char-long-context} compares speedups across
different tree sizes on two representative datasets: NQA (short context and
output) and CNN (long context and output).
As shown in \figref{char-long-context}(a), CNN exhibits higher speedups that
vary more widely with tree size than those of NQA, despite its lower acceptance
length (3.08 vs.\ 3.22).
We attribute this to two factors.

First, as shown in \figref{char-long-context}(b), attention accounts for a
larger fraction of the total execution time in CNN due to its longer context
(1043 tokens vs.\ 281 for NQA).
This makes overall inference more bandwidth-bound and reduces the relative cost
of the additional computation that SD introduces in the linear layers.

Second, the average batch size for CNN (18.0) is larger than that for NQA (3.4)
because longer outputs cause requests to linger in the serving system.
An optimal tree enables earlier request completion, which directly reduces the
batch size and thus per-iteration execution time.
For instance, as shown in \figref{char-long-context}(c), SD with an optimal tree
reduces the average batch size of CNN from 18.0 to 4.0, resulting in a
1.7$\times$ reduction in per-iteration execution time.

In sum, long-context workloads widen the gap between optimal and suboptimal tree
configurations, thereby making adaptive tree construction increasingly
consequential.

\putsec{design}{\name{}: A Plug-in Component for Adaptive Speculative Decoding}

We propose \name{}, a plug-in component for LLM serving systems that dynamically
controls tree size and depth based on the request rate and per-request
characteristics.
\name{} selects the tree configuration that maximizes estimated throughput,
which captures the trade-off between execution cost and acceptance length.

\putssec{design-optimization-obj}{Optimization Objective}
We define \textit{per-iteration throughput} as the total number of accepted
tokens divided by the batch execution time,
following~\cite{liu:dan24, wu:zho25}.
The per-iteration throughput, denoted as $R$, is expressed as a function of two
variables related to SD, tree size $k_s$ and tree depth $d$, along with two
factors that depend on the serving conditions, batch size $b$, determined by the
request rate of the system, and requests within a batch $r_{1:b}$:
\begin{equation}
\label{eqn:performance-modeling}
\begin{aligned}
  R &= \frac{ \sum_{j=1}^{b}{\tau(r_j)}}
  {d \cdot T(k_e, b, M_D) + T(k_s, b, M_T)},
\end{aligned}
\end{equation}
where $k_e$ denotes the tree expansion width described in \ssecref{tree-sd}
(fixed at 10 following ~\cite{li:wei24-eagle2}), $\tau(r_j)$ denotes the
acceptance length for the $j$-th request in the batch, $M_T$ and $M_D$ refer to
the target and draft models, respectively, and $T$ denotes the execution time
for a given model.
Given the batch size and requests at each iteration, we aim to determine the
optimal values of $k_s$ and $d$ that maximize the per-iteration throughput $R$,
as expressed below.
\begin{equation}
\label{eqn:optimization-obj}
\begin{aligned}
  k_s, d &= \arg\max_{k_s, d} R{(k_s, d, b, r_{1:b})} \\
\end{aligned}
\end{equation}
In the following subsections, we describe how to estimate the values of the
functions in \eqnref{performance-modeling}, namely $\tau$ and $T$.

\putssec{design-acceptance-length}{Estimating Acceptance Length}
Since the acceptance length $\tau(r_j)$ is unknown prior to the target model's
verification, we use an estimated value $\hat{\tau}(r_j)$ to predict the
acceptance length of a given token tree, $Tr$.
We begin with the expectation based on the draft model's
logits~\cite{li:wei24-eagle2}, as follows:

\begin{equation}
\label{eqn:estimation_accept_length}
\tau(\mm{r}{j}) \approx \hat{\tau}(r_j)=\sum_{v \in r_j} p_{v}, \text{where  }  p_{v}=\prod_{u \in path(v)}P(u|c_u).
\end{equation}
$v$ denotes each node in the draft tree, and $p_{v}$ represents the probability
of acceptance for the node $v$ based on the draft model's logits for the request
$r_j$.
Node $u$ refers to a token residing in a draft path between the root and node
$v$ ($path(v)$), and $c_u$ corresponds to the preceding tokens of node $u$.

Since the small draft model does not exactly match the target model's
probability distribution, the draft model can be overconfident in its
predictions.
This can lead to an overestimated acceptance length
(\eqnref{estimation_accept_length}) and trigger wasteful executions of the draft
model.

To mitigate this overestimation, we apply temperature scaling~\cite{guo:ple17}
as a calibration technique for the draft model's logits before computing softmax
probabilities to smooth the probability distribution when calculating
\eqnref{estimation_accept_length}.

While temperature scaling affects the entire distribution, reducing
overestimated probabilities is more critical for system efficiency.
This is because such overestimation directly leads to unnecessary executions of
the draft model, whereas underestimation only results in more conservative
decisions, without wasting any computation, since the undrafted tokens can still
be drafted in the next iteration.

\begin{figure}[t]
    {
        \includegraphics[width=\textwidth]{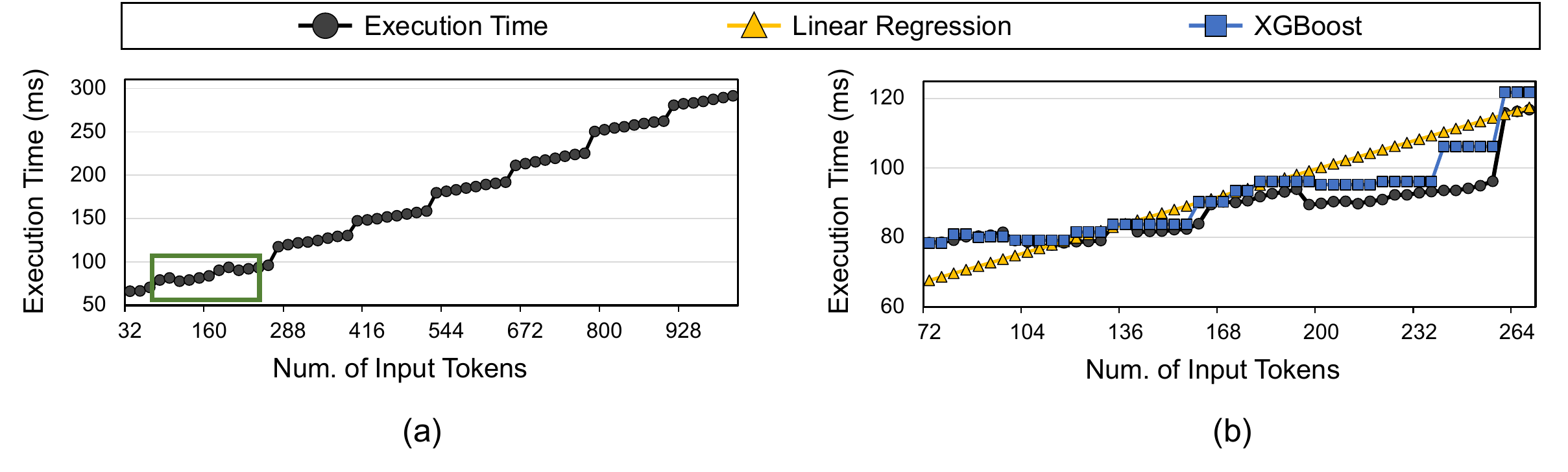}
        \vspace{-0.25in}
    }
    \caption{(a) Model execution time as a function of the number of input
      tokens for Qwen3-32B on a Pro 6000 Blackwell GPU.
      (b) Comparison of execution time predictions using linear regression and
      XGBoost against the measured execution time shown in (a).}
    \vspace{-0.15in}
    \label{fig:design-model-exec-time-blackwell}
\end{figure}

\putssec{design-execution-time}{Estimating Execution Time using XGBoost}

Since we cannot determine the exact execution time $T$ in
\eqnref{performance-modeling}, we design an execution time estimation function,
$\hat{T}$, for the model $M$ and GPU of the serving system.
This function, $\hat{T}$, takes three inputs that influence arithmetic intensity
and execution time: batch size $b$, context length $n_{context}$ (i.e., the
number of tokens of the KV cache), and input token length $n_{token}$.
Note that $n_{token}$ is the tree size $k_s$ for the target model and the tree
expansion width $k_e$ for the draft model.
The estimated execution times for the target model $M_T$ and the draft model
$M_D$ are expressed as follows.

\vspace{-10pt}
\begin{equation}
\label{eqn:estim-exec-time-target}
T(k_s, b, M_T) \approx \hat{T}_{M_T}(b, n_{context}, k_s),
\end{equation}

\vspace{-15pt}

\begin{equation}
\label{eqn:estim-exec-time-draft}
T(k_e, b, M_D) \approx \hat{T}_{M_D}(b, n_{context}, k_e)
\end{equation}

\myparagraph{Nonlinear Behavior of Model Execution Time.}
The model execution time for a single iteration can be approximated as a linear
function~\cite{liu:dan24} of the total number of context tokens within a batch,
$N_{context}=\sum_{j=1}^b{n_{context,j}}$, and the total number of input tokens
within a batch, $N_{token}=\sum_{j=1}^b{n_{token,j}}$, based on the model's
arithmetic intensity.
However, the actual execution time exhibits nonlinear behavior with respect to
$N_{token}$, as the linear operator gradually shifts from being memory
bandwidth-bound to compute-bound as arithmetic intensity increases.

\figref{design-model-exec-time-blackwell}(a) shows the target model's execution
time as a function of the number of input tokens.
When the number of tokens is small, execution time remains nearly constant, as
loading the model weights dominates the small amount of computation, keeping the
operation memory bandwidth-bound.
As the number of tokens increases, however, the execution time scales
approximately linearly with the number of tokens, because the increased
computation translates directly into additional execution latency.
Due to this \textit{nonlinear} behavior, execution time cannot be accurately
modeled using a linear equation, as shown by the yellow line in
\figref{design-model-exec-time-blackwell}(b).

\myparagraph{XGBoost as a Predictor for Model Execution Time.}
To handle this nonlinearity, we employ
\textbf{\textit{XGBoost}}~\cite{che:gue16}, an ensemble of decision trees, to
predict execution time.
The model takes as input a 3-dimensional vector: batch size $b$, maximum context
length within a batch $n_{context,max}$, and the total number of input tokens
$N_{token}=\sum_{j=1}^b{n_{token,j}}$.
Each decision tree produces a prediction by traversing branches based on binary
conditions at each node.
These outputs from all decision trees are then aggregated to form the final
prediction of execution time, which allows the model to capture the nonlinear
trends as the number of input tokens varies.

XGBoost outperforms linear regression in predicting execution time, particularly
when compute resources are underutilized with low arithmetic intensity, which is
a typical scenario in speculative decoding.
As shown in \figref{design-model-exec-time-blackwell}(b), the blue line
(XGBoost) closely follows the flat trends in bandwidth-bound regions, while the
yellow line (linear regression) always predicts a linear trend.

\begin{figure}[t]
    {
        \includegraphics[width=\textwidth]{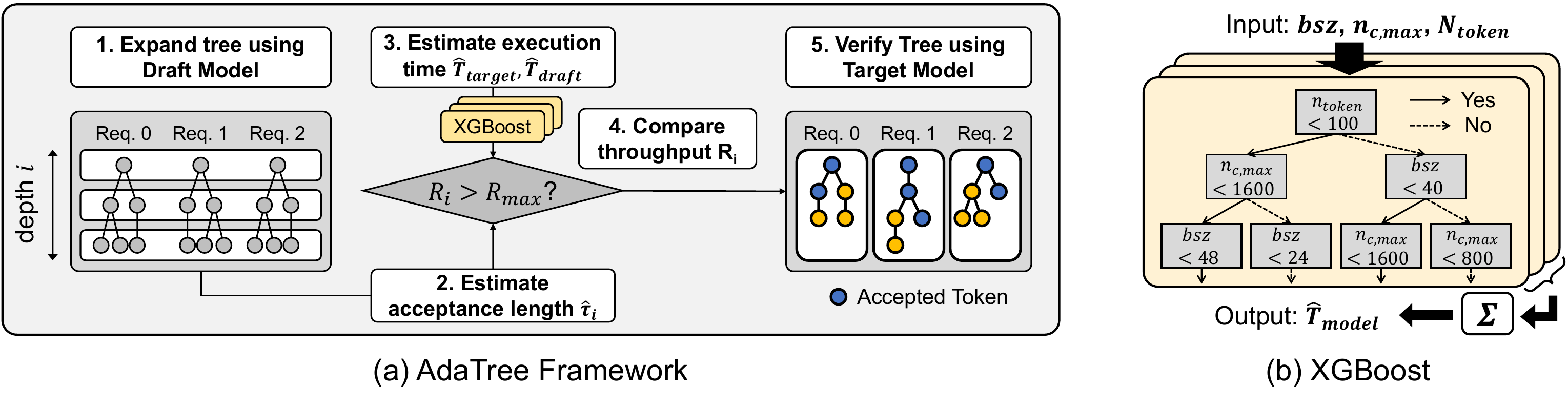}
    }
    \vspace{-0.25in}
    \caption{(a) Overview of our framework, \name{}.
      (b) XGBoost that captures nonlinear model execution time.}
    \vspace{-0.15in}
    \label{fig:design-framework}
\end{figure}

\putssec{design-framework}{Integration of Performance Estimator into the Generation Pipeline}

\figref{design-framework}(a) shows the \name{} framework, in which the
estimation methods described in \ssecref{design-acceptance-length} and
\ssecref{design-execution-time} are integrated into the generative LLM inference
pipeline.

\myparagraph{Execution Flow of \name{}.}
\name{} determines the optimal tree structure through the following iterative
process:

\textbf{Step 1. Tree expansion}: The draft model runs once to create an initial
tree of depth 1 (i.e., $i$=1).

\textbf{Step 2. Estimation of acceptance length}: \name{} estimates the total
acceptance length of the batch, $\sum_{j=1}^{b}{\hat{\tau}(r_j)}$, for varying
candidate tree sizes using the draft model's logits.

\textbf{Step 3. Estimation of execution time}: \name{} predicts execution times
for both the target and draft models, $\hat{T}_{M_T}$ and $\hat{T}_{M_D}$,
across candidate sizes using XGBoost (\figref{design-framework}(b)).

\textbf{Step 4. Throughput comparison}: Based on the estimates from
\textbf{Steps 2} and \textbf{3}, \name{} calculates per-iteration throughput at
the current depth using \eqnref{performance-modeling}, and identifies $R_1$, the
maximum throughput across the candidate sizes.
For the initial depth, $R_{max}$ is initialized to $-\infty$ and is immediately
updated to $R_1$.

Following the initial iteration, the draft model runs again (\textbf{Step 1}) to
expand the tree by one level (i.e., $i$=2).
\name{} estimates the acceptance length and execution times in the same manner
as at depth 1 (\textbf{Steps 2} and \textbf{3}), and determines the maximum
per-iteration throughput $R_2$ (\textbf{Step 4}).
If $R_2$ exceeds $R_{max}$, which indicates that a deeper tree is likely to
improve throughput further, $R_{max}$ is updated to $R_2$, and the same process
(\textbf{Steps 1}--\textbf{4}) continues for the next depth (i.e., 3).
Otherwise, tree expansion is immediately terminated, which effectively reduces
wasteful execution of the draft model.

\textbf{Step 5. Tree verification}: The final trees are constructed using the
size that yielded the maximum throughput $R_i$ during the last iteration of the
draft model.
Trees are then verified in parallel by the target model, and the entire process
(\textbf{Steps 1}--\textbf{5}) is repeated using the newly accepted tokens.

We provide a walk-through example in \ssecref{appendix-example}.

\putsec{evaluation}{Evaluation}

\putssec{setting}{Setting}

\myparagraph{Models.}
We use Llama-3.1-8B-Instruct~\cite{gra:dub24} as the target model and its draft
model trained with EAGLE-3~\cite{li:wei25}, a state-of-the-art method for
training draft models.
Results on additional models are provided in \ssecref{appendix-model} and
\ssecref{appendix-eagle}.
All model weights are obtained from Hugging Face~\cite{wol:deb20}.

\myparagraph{Hardware Platforms.}
We evaluate \name{} on a recent high-end GPU, the NVIDIA RTX Pro 6000 Blackwell.
To demonstrate generalizability, we also provide further evaluations on a
consumer-grade GPU, the RTX 3090, in \ssecref{appendix-consumer-GPU}.

\myparagraph{Datasets.}
We evaluate on Spec-Bench~\cite{xia:yan24}, a benchmark for speculative decoding
that includes six datasets, each representing a distinct language modeling task:
MT-Bench (multi-turn conversation), WMT (German-to-English translation), CNN
(summarization), NQA (question answering), GSM8K (mathematical reasoning), and
DPR (retrieval-augmented generation).
We additionally evaluate on HumanEval~\cite{che:two21}, a benchmark for code
completion, and ShareGPT~\cite{sharegpt2023}, which consists of prompts from
user-shared conversations paired with GPT-4-generated
responses~\cite{wan:che24}.
Request arrivals follow a Poisson process following
~\cite{yu:jeo22, kwo:li23, agr:ked24}, with the maximum generation length
limited to 512.

\myparagraph{Baseline Designs.}
We compare \name{} to AD, SpS (speculative decoding using a static
chain)~\cite{che:bor23}, and EAGLE-3~\cite{li:wei25}.
For EAGLE-3, we use a tree size of 60 and a depth of 8, as specified in
~\cite{li:wei25}.
For SpS, we set the chain length to 8 to match the depth of EAGLE-3.

\myparagraph{Implementation.}
We extend the official codebase of EAGLE-3 to support online serving for
tree-based SD, where requests are dynamically added to and removed from the
batch at each iteration~\cite{yu:jeo22}.
Tree attention for batched inference is implemented using
Triton~\cite{til:kun19}.
For acceptance length estimation, we apply temperature scaling with a value of
1.2 to the logits of the draft model.

\putssec{main-results}{Main Results}
We evaluate performance by measuring the average end-to-end execution time for
requests and report it as a speedup over AD.
To show that \name{} consistently outperforms SD with static configurations
under fluctuating serving conditions, we vary the request rate and
characteristics.
For each condition, we also analyze the main source of the performance gain.

\putsssec{eval-request-rate}{Performance Comparison with Varying Request Rates}

\begin{figure}[t]
    {
        \begin{center}
            \vspace{-0.10in}
            \includegraphics[width=\textwidth]{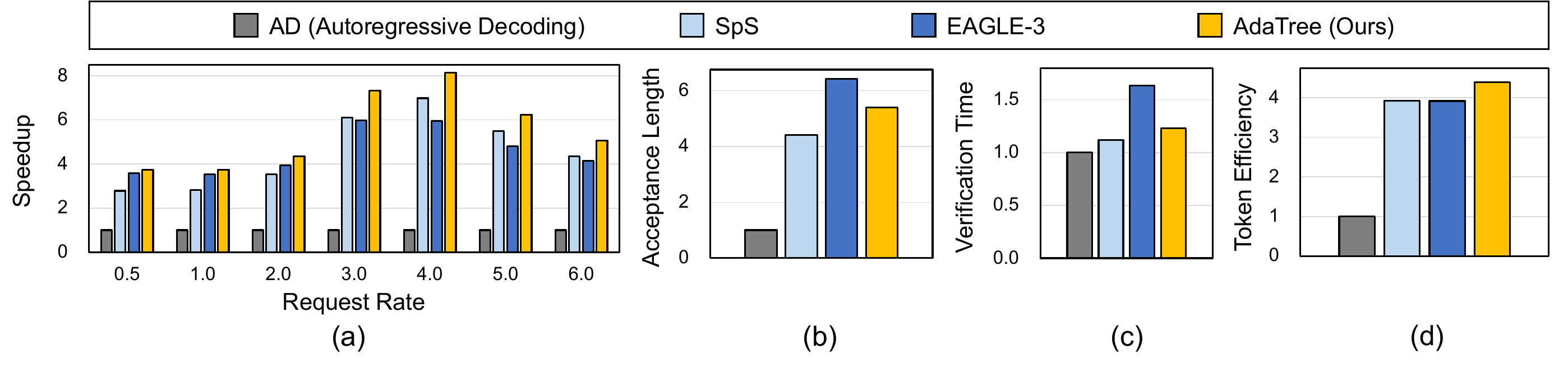}
        \end{center}
        \vspace{-0.20in}
    }
    \caption{Performance comparison with varying request rates using GSM8K
      dataset on RTX Pro 6000 Blackwell.
      (a) Speedup of SpS, EAGLE-3, and \name{} over AD.
      (b) Average acceptance length.
      (c) Normalized verification time to that of AD at a batch size of 3.
      (d) Token efficiency, defined as the average acceptance length divided by
      the verification time.}
    \vspace{-0.10in}
    \label{fig:eval-llama-blackwell-eagle3-req-rate}
\end{figure}

\figref{eval-llama-blackwell-eagle3-req-rate} compares \name{} with the
baselines.
As shown in \figref{eval-llama-blackwell-eagle3-req-rate}(a), \name{}
outperforms SpS and EAGLE-3 across all request rates, achieving up to a
1.2$\times$ speedup over the better of the two baselines at 3.0~requests/s.
This gain stems mainly from \textit{dynamically adjusting the tree size} to the
request rate and the resulting batch size.
\name{} selects a tree size suited to the current batch size, preventing
verification from becoming heavily compute-bound while maintaining a higher
acceptance length than SpS.

To identify the source of the performance gain, we report acceptance length,
verification time, and \textit{token efficiency}, which is defined as the
acceptance length per unit verification time, in
Figures~\ref{fig:eval-llama-blackwell-eagle3-req-rate}(b)-(d) at a request rate
of 2.0.
Verification time is measured at a batch size of 3, which approximates the
average batch sizes observed under SpS, EAGLE-3, and \name{} (3.1, 2.6, and 2.6,
respectively).

As shown in \figref{eval-llama-blackwell-eagle3-req-rate}(b), \name{} achieves a
higher acceptance length (5.4) than SpS (4.4), preserving the primary benefit of
tree-based drafting over chain-based drafting.
Meanwhile, by constructing a smaller tree (size 32) than EAGLE-3 (size 60),
\name{} reduces verification time by 1.3$\times$, as shown in
\figref{eval-llama-blackwell-eagle3-req-rate}(c).
Consequently, as shown in \figref{eval-llama-blackwell-eagle3-req-rate}(d),
\name{} achieves the highest token efficiency by identifying an optimal balance
between acceptance length and verification overhead.

Note that the end-to-end speedup in
\figref{eval-llama-blackwell-eagle3-req-rate}(a) exceeds the token efficiency
gain in \figref{eval-llama-blackwell-eagle3-req-rate}(d), as the former is
evaluated in a dynamic serving environment where batch sizes fluctuate across
iterations, and early-exiting requests progressively reduce the batch size,
further amplifying \name{}'s advantage.

\putsssec{eval-request-char}{Performance Comparison with Varying Request Characteristics}
\begin{figure}[t]
    {
        \begin{center}
            \includegraphics[width=\textwidth]{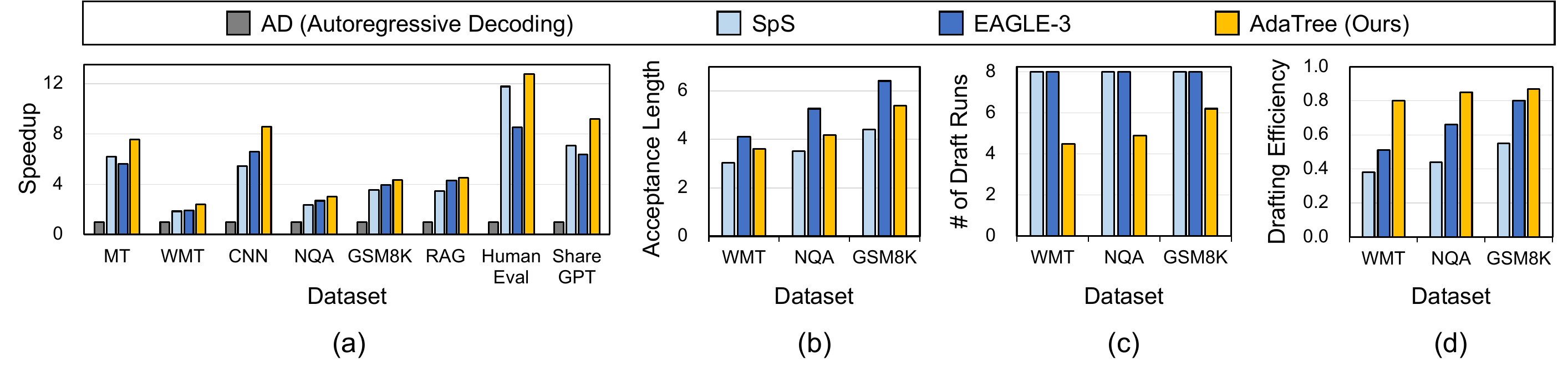}
        \end{center}
        \vspace{-0.15in}
    }
    \caption{Performance comparison with varying datasets on RTX Pro 6000
      Blackwell at request rate of 2.0.
      (a) Speedup of SpS, EAGLE-3, and \name{} over AD.
      (b) Average acceptance length.
      (c) Average number of draft model runs.
      (d) Drafting efficiency, defined as the average acceptance length divided
      by the number of draft model runs.}
    \vspace{-0.15in}
    \label{fig:eval-llama-blackwell-eagle3-req-char}
\end{figure}

As shown in \figref{eval-llama-blackwell-eagle3-req-char}(a), \name{}
outperforms SpS and EAGLE-3 across all datasets, achieving up to 1.3$\times$
speedup over the better of the two baselines on CNN, with an average of
1.26$\times$ speedup across all datasets.
The key source of speedup is the \textit{dynamic adjustment of tree depth} in
response to diverse request characteristics.
\name{} achieves this by estimating the throughput at every depth and
terminating tree expansion as soon as it determines that constructing a deeper
tree would be wasteful, avoiding unnecessary draft model runs.

To identify the source of this performance gain, we report acceptance length,
the number of draft model runs, and \textit{drafting efficiency}, which is
defined as acceptance length per number of draft model runs, in
\figref{eval-llama-blackwell-eagle3-req-char}(b)-(d).
We select three datasets representing low, mid, and high acceptance lengths:
WMT, NQA, and GSM8K, respectively.

SpS and EAGLE-3 always run the draft model a fixed number of times, which causes
wasteful executions when the average acceptance length is relatively low.
For example, as shown in \figref{eval-llama-blackwell-eagle3-req-char}(b), WMT
exhibits an acceptance length of 4.1 with EAGLE-3, implying that nearly half of
draft model runs (i.e., 8 times) are wasteful.
For such requests, \name{} constructs a shallower tree, executing the draft
model only 4.5 times, while achieving an acceptance length of 3.6, as shown in
\figref{eval-llama-blackwell-eagle3-req-char}(b)-(c).
In contrast, for requests with a high acceptance length, such as GSM8K (6.4 with
EAGLE-3), \name{} constructs a deeper tree (depth of 6.2) to fully utilize its
prediction capability.
Consequently, as shown in \figref{eval-llama-blackwell-eagle3-req-char}(d),
\name{} achieves the highest drafting efficiency across diverse request
characteristics by adaptively terminating tree expansion at the optimal depth.

\myparagraph{Requests with Long Context and Output.}
As shown in \figref{eval-llama-blackwell-eagle3-req-char}(a), MT, CNN, and
ShareGPT exhibit higher speedups compared to other datasets, even though their
acceptance lengths are not the highest.
This is because their context and output lengths are long, which lead to larger
batch sizes under the same request rate.
Consequently, the performance gain of \name{}, which enables earlier exit of
requests, is maximized because the reduced batch size substantially reduces the
per-iteration latency, as discussed in \ssecref{motiv-long-context}.
For example, \name{} reduces the average batch size of MT from 28.6 to 5.9.

\myparagraph{Serving a Realistic Request Trace.}
We evaluate \name{} on the Azure LLM inference trace~\cite{azu23}, which
provides real-world request arrival times along with input and output lengths.
Because the trace omits prompt contents for privacy, we sample each prompt from
the PG-19 dataset~\cite{rae:pot20} to match the input length of the trace.
We replay the first 5K requests and report speedup over AD, along with P50, P90,
and P99 latency.

\begin{table}[t]
    \centering
    \caption{Speedup and tail latency on the first 5K requests of the Azure LLM
      inference trace.}
    \label{tab:eval-azure-trace}
    \small
    \begin{tabular}{lcccc}
        \toprule
        Scheme & Speedup & P50 & P90 & P99 \\
        \midrule
        AD       & 1.00$\times$ & 63.9s & 75.2s & 109.6s \\
        SpS      & 1.21$\times$ & 48.6s & 70.5s & 97.0s \\
        EAGLE-3  & 0.73$\times$ & 68.6s & 101.5s & 139.0s \\
        \textbf{AdaTree} & \textbf{1.60}$\times$ & \textbf{43.2s} & \textbf{57.8s} & \textbf{80.5s} \\
        \bottomrule
    \end{tabular}
\end{table}

As shown in \tabref{eval-azure-trace}, \name{} outperforms static SD baselines
on the real-world trace, which exhibits greater fluctuation in request rate
(standard deviation of 2.4, compared to 2.2 of a Poisson distribution) and
higher variance in output length (standard deviation of 168 vs. 104 for
Spec-Bench).
Importantly, \name{} does not degrade P50, P90, or P99 latency demonstrating
that it does not penalize the performance of individual requests.

Note that the system heavily shifts into a compute-bound regime for this trace
because the long average output length (288) causes requests to linger longer in
the system and increases the average batch size (e.g., 59 for AD) under the
average request rate of 4.9.
This is why EAGLE-3 with a static tree even performs worse than AD.
We emphasize that AdaTree effectively survives and outperforms baselines in this
extreme scenario (keeping the 1.32$\times$ speedup over the better of the
baselines) by adapting to the serving scenario, constructing a much smaller tree
by constraining the size and depth.
We note that under such compute-bound conditions, hardware with a higher
TFLOPS-to-memory-bandwidth ratio would be required to increase the absolute
speedup of SD.

\putsec{conclusion}{Conclusion and Limitation}

Speculative decoding is a widely adopted technique for mitigating the memory
bandwidth-bound nature of autoregressive LLM inference.
When serving LLMs with SD, however, a statically configured token tree leads to
suboptimal performance.
In this work, we focus on two key parameters of the token tree that primarily
determine SD performance: tree size and depth.
Building on an analysis of how their optimal values vary with request rate and
request characteristics, we propose \name{}, a plug-in component that enhances
SD in LLM serving frameworks.
By estimating model execution time and acceptance length, \name{} dynamically
adjusts tree size and depth, consistently delivering higher performance than SD
with static chains and trees across diverse serving conditions.

We evaluate \name{} with SD using a separate autoregressive transformer draft
model, as it achieves state-of-the-art acceptance rates.
Future work could extend \name{} to self-drafting models, non-transformer models
(e.g., statistical models), or recently proposed block diffusion drafters that
generate multiple draft tokens in parallel.
For example, such parallel drafters may require revisiting how we model the
execution time of the draft model, as they generate logits for multiple draft
positions at once, unlike sequential autoregressive drafters.

\newpage
\begin{ack}
We are grateful to the anonymous reviewers for their constructive comments and
valuable feedback.
This work was supported by the Institute for Information \& Communications
Technology Planning \& Evaluation (IITP) grants funded by the Korean government
(MSIT) (No.
RS-2025-02214652, No. RS-2026-25524173, No. RS-2022-00156295).
Jaewoong Sim is the corresponding author.
\end{ack}

\bibliographystyle{plain}
\bibliography{refs}

\newpage
\appendix
\putsec{appendix}{Appendix}

\putssec{appendix-example}{A Walk-Through Example of \name{}}
\begin{figure}[htbp]
    {
        \begin{center}
            \vspace{-10pt}
            \includegraphics[width=\textwidth]{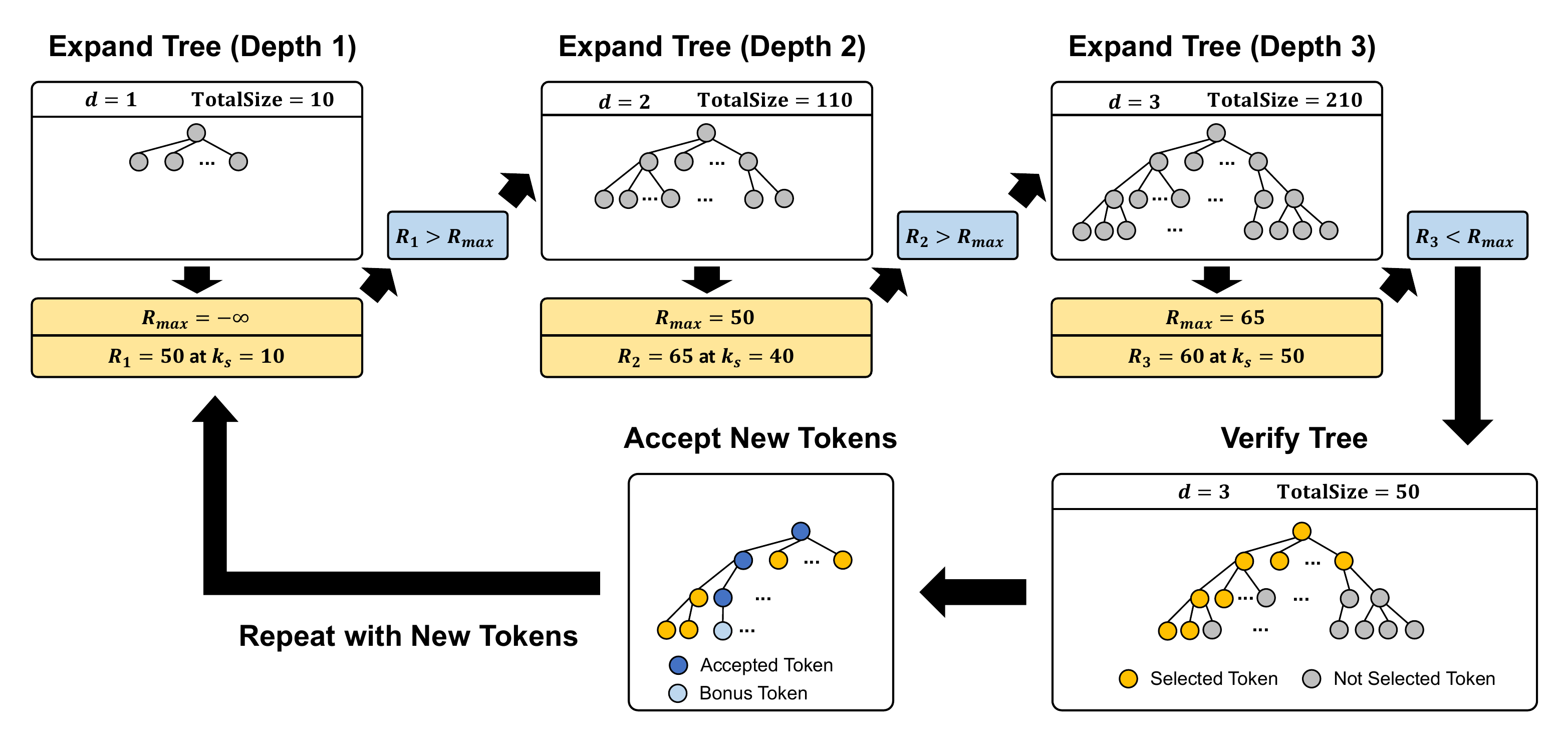}
            \vspace{-15pt}
        \end{center}
    }
    \caption{A walk-through example in which \name{} selects an optimal tree of
      size 50 and depth 3.}
    \label{fig:appendix-A.1-example}
\end{figure}

In \figref{appendix-A.1-example}, we provide a walk-through example in which
\name{} selects an optimal tree of size 50 and depth 3.

The draft model constructs an initial tree with depth 1 and size 10.
At depth 1, the maximum throughput is estimated as 50 at size 10 ($R_1 =$ 50).
Since $R_{max}$ is initialized as negative infinity, we continue to expand the
tree.

The draft model runs again and then expands the tree to depth 2, with a total
size of 110 (10 original nodes + 100 new child nodes).
At depth 2, the maximum throughput is estimated as 65 at size 40 ($R_2 =$ 65).
Since this is higher than that at depth 1 ($R_2 > R_1$), we continue to expand
the tree.

The draft model runs again and then expands the tree to depth 3, with a total
size of 210 (110 original nodes + 100 new child nodes).
At depth 3, the maximum throughput is estimated as 60 at size 50 ($R_3 =$ 60).
Since this is lower than that at depth 2 ($R_3 < R_2$), we terminate tree
expansion.

We construct the final tree of depth 3 and size 50, which yields the maximum
throughput at the final depth.
Note that the tree of size 40 at depth 2 ($R_2 =$ 65) is no longer attainable at
this point: the draft model has already run three times, and this drafting cost
is incurred regardless of which tree is selected.
Therefore, among the trees available after the third draft model execution, we
select the one with the maximum estimated throughput.
The target model verifies it in parallel and accepts tokens that match its
output.

The process above repeats using the newly accepted tokens.

\putssec{appendix-model}{Extension to Other Target Models}
To demonstrate the generalizability of our framework to diverse model types and
scales, we additionally evaluate the Qwen3~\cite{yan:li25} series, specifically
Qwen3-8B and Qwen3-32B.
As discussed in \ssecref{design-execution-time}, we train separate XGBoost
models to predict the execution time for each model.
Using the trained XGBoost models, our framework supports new target models
without tuning any other thresholds.
We evaluate \name{} using the Qwen3 series as the target model and the draft
model from EAGLE-3, similar to the methodology in \ssecref{main-results}.

\myparagraph{Performance Comparison to Llama-3.1-8B-Instruct}

\begin{figure}[htbp]
    {
        \begin{center}
            \includegraphics[width=\textwidth]{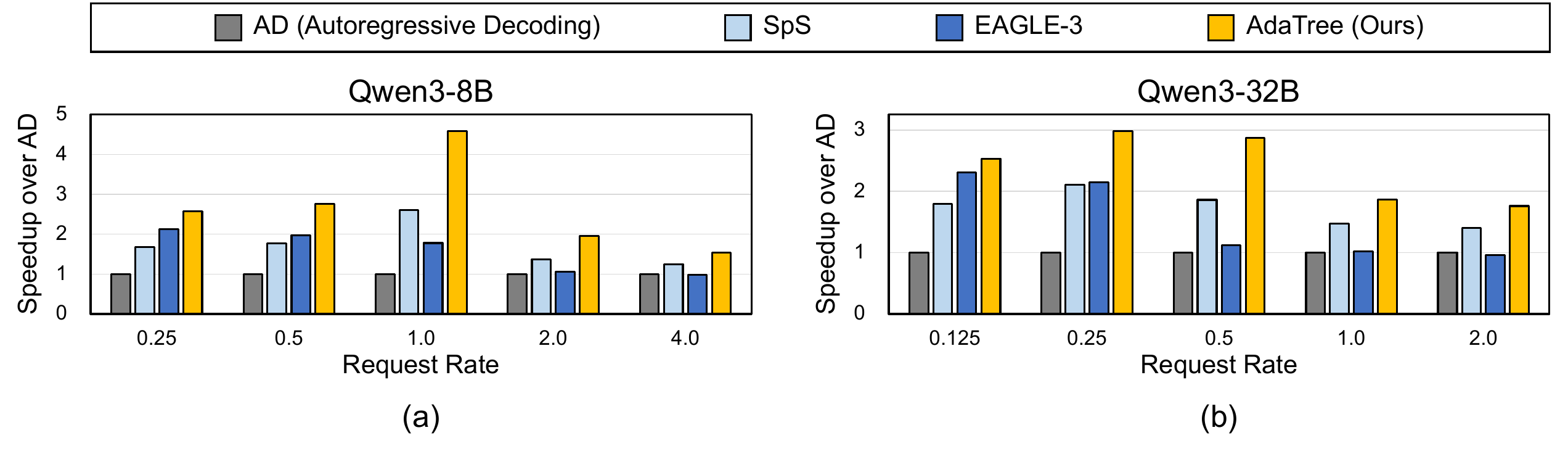}
        \end{center}
        \vspace{-15pt}
    }
    \caption{Speedup of SpS, EAGLE-3, and \name{} over AD with varying request
      rates for (a) Qwen3-8B and (b) 32B, using EAGLE-3 and GSM8K dataset on Pro
      6000 Blackwell.
    }
    \label{fig:appendix-qwen3-blackwell-eagle3-req-rate}
\end{figure}

\begin{figure}[htbp]
    {
        \begin{center}
            \includegraphics[width=\textwidth]{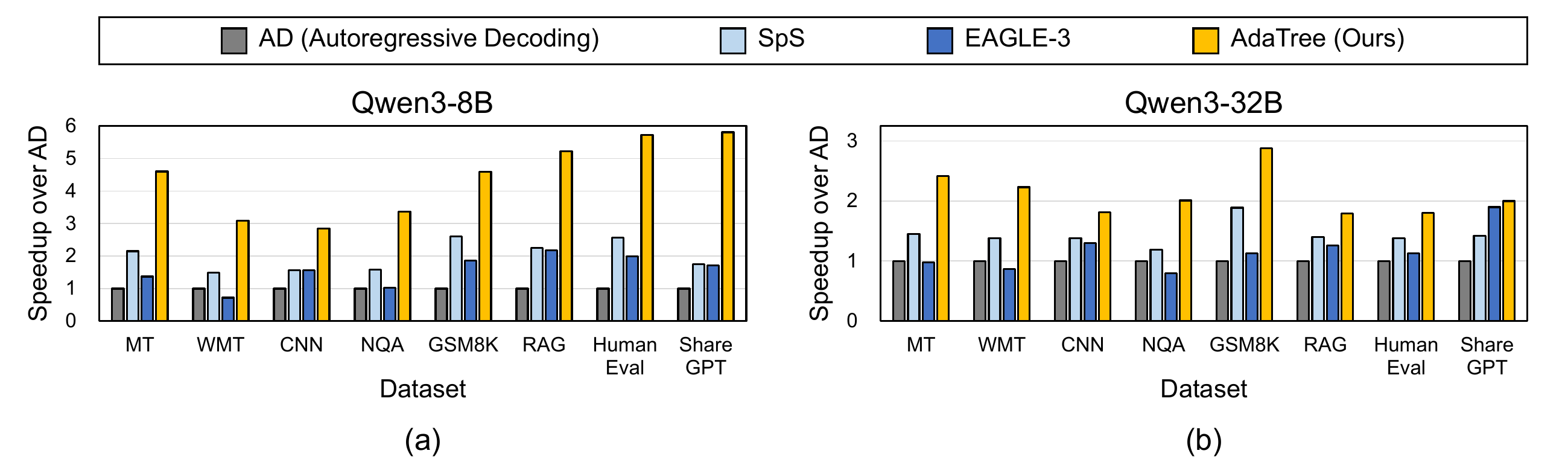}
        \end{center}
        \vspace{-15pt}
    }
    \caption{Speedup of SpS, EAGLE-3, and \name{} over AD with varying datasets
      for (a) Qwen3-8B at 1.0 requests/s and (b) 32B at 0.5 requests/s, using
      EAGLE-3 on Pro 6000 Blackwell.
    }
    \label{fig:appendix-qwen3-blackwell-eagle3-req-char}
\end{figure}

As shown in \figref{appendix-qwen3-blackwell-eagle3-req-rate}(a), (b), and
\figref{appendix-qwen3-blackwell-eagle3-req-char}(a), (b), \name{} consistently
outperforms SpS and EAGLE-3 across varying request rates and datasets for both
Qwen3-8B and Qwen3-32B, demonstrating that \name{} is generalizable and scalable
across different model families and sizes.

When compared to Llama-3.1-8B-Instruct, the target model used in our main
evaluation in \ssecref{main-results}, the speedup of SpS and EAGLE-3 over AD is
generally lower for Qwen3.
This is because Qwen3 exhibits a lower acceptance length for the same dataset.
For example, for GSM8K, the acceptance length for Llama-3.1-8B-Instruct is 6.4,
whereas those for Qwen3-8B and 32B are 3.9 and 3.4, respectively.

On the other hand, the relative performance gain of \name{} over the baselines
(SpS and EAGLE-3) is larger for Qwen3.
For example, \name{} achieves up to 1.8$\times$ and 1.5$\times$ speedup over the
better of SpS and EAGLE-3 for Qwen3-8B and 32B, respectively, while achieving
1.2$\times$ for Llama-3.1-8B-Instruct.
This is because Qwen3, which is trained as a thinking model~\cite{yan:li25},
produces longer outputs due to the intermediate thinking content, which
increases the average batch size, as discussed in \ssecref{motiv-long-context}.
Consequently, the effect of early request exit enabled by \name{} is maximized,
as the optimal tree selection more significantly reduces the batch size and
per-iteration execution time.

\putssec{appendix-eagle}{Extension to Other Draft Models}
There are other methods for training the draft model, such as
EAGLE~\cite{li:wei24-eagle1}.
While our main evaluation employs EAGLE-3~\cite{li:wei25}, which achieves
state-of-the-art performance, we additionally provide experimental results with
EAGLE to demonstrate the scalability of our framework across different draft
model versions.
Our framework can be seamlessly adapted to a new draft model by simply training
a lightweight XGBoost for it, which takes only a few minutes.
We evaluate our framework using the draft model trained by EAGLE, similar to the
methodology in \ssecref{main-results}, with EAGLE-2's tree construction method
for fair comparison.

\myparagraph{Performance Comparison with Varying Request Rates}

\begin{figure}[htbp]
    {
        \begin{center}
            \includegraphics[width=\textwidth]{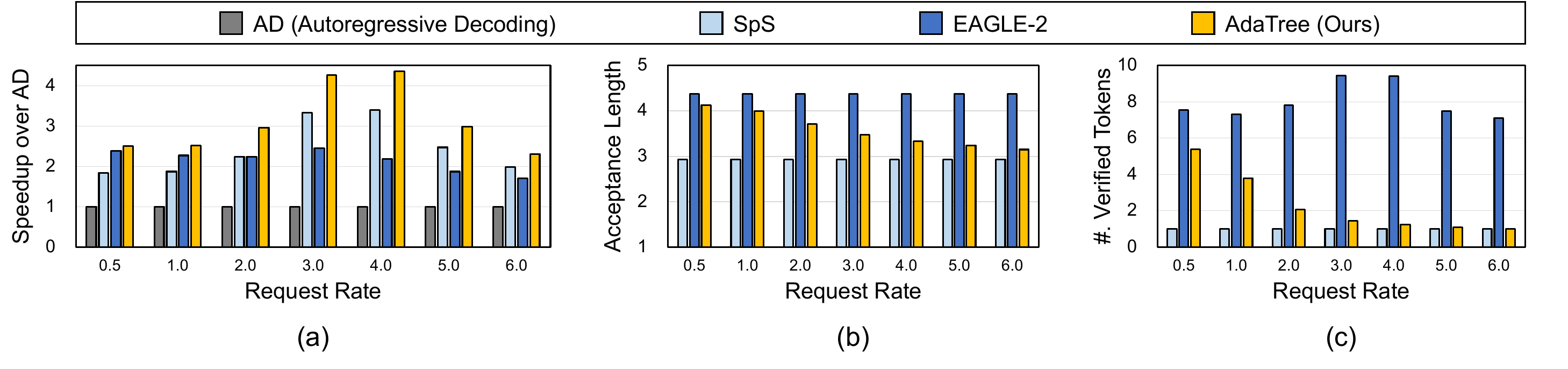}
        \end{center}
        \vspace{-15pt}
    }
    \caption{Performance comparison with varying request rates using the draft
      model from EAGLE and GSM8K dataset on Pro 6000 Blackwell.
      (a) Speedup of SpS, EAGLE-2, and \name{} over AD.
      (b) Average acceptance length.
      (c) Average number of tokens processed per target model iteration,
      normalized to that of SpS.}
    \label{fig:appendix-llama-blackwell-eagle2-req-rate}
\end{figure}

As shown in \figref{appendix-llama-blackwell-eagle2-req-rate}(a), the speedup of
SpS and EAGLE-2 over AD is generally lower than that achieved with EAGLE-3
(\figref{eval-llama-blackwell-eagle3-req-rate}(a)).
This is because the draft model trained by EAGLE-3 predicts target model tokens
more accurately than EAGLE, yielding a higher average acceptance length (e.g.,
6.41 vs 4.38 for the same dataset, GSM8K), as shown in
\figref{eval-llama-blackwell-eagle3-req-rate}(b) and
\figref{appendix-llama-blackwell-eagle2-req-rate}(b).

Despite the variations in draft model performance, \name{} consistently
outperforms SpS and EAGLE-2, as shown in
\figref{appendix-llama-blackwell-eagle2-req-rate}(a).
The primary source of performance gain is its dynamic adjustment of tree size
according to the request rate, consistent with the behavior of \name{} observed
with EAGLE-3.
Trends in the number of verified tokens by \name{}
(\figref{appendix-llama-blackwell-eagle2-req-rate}(c)) remain similar to those
observed with EAGLE-3, because the computational overhead from verification is
determined by the target model, regardless of the choice of the draft model.

\myparagraph{Performance Comparison with Varying Request Characteristics}

\begin{figure}[htbp]
    {
        \begin{center}
            \includegraphics[width=\textwidth]{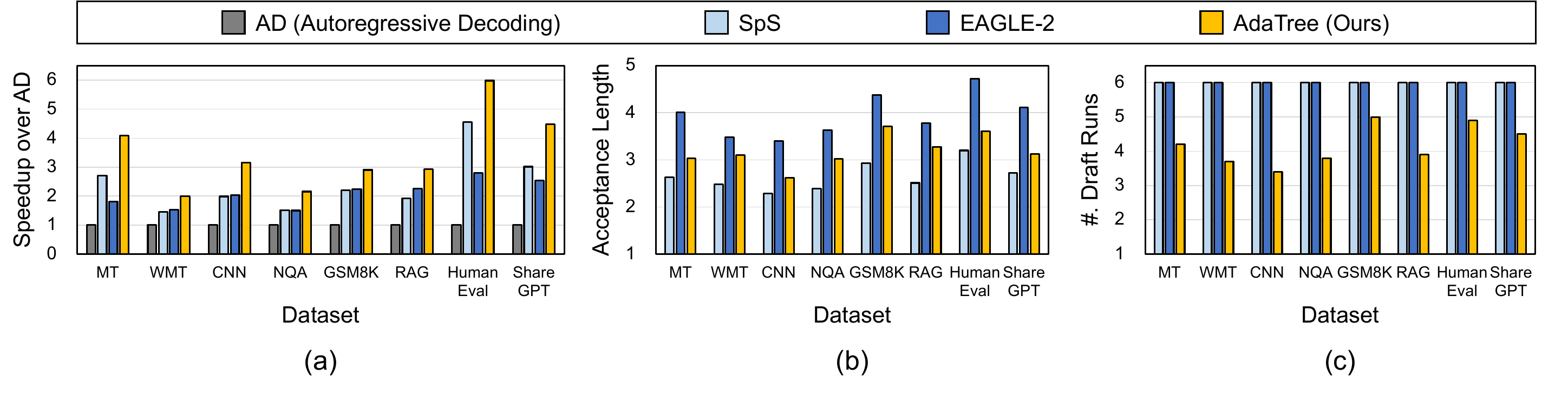}
        \end{center}
        \vspace{-15pt}
    }
    \caption{Performance comparison with varying datasets using the draft model
      from EAGLE at 2.0 requests/s on Pro 6000 Blackwell.
      (a) Speedup of SpS, EAGLE-2, and \name{} over AD.
      (b) Average acceptance length.
      (c) Average number of draft model runs.}
    \label{fig:appendix-llama-blackwell-eagle2-req-char}
\end{figure}

As shown in \figref{appendix-llama-blackwell-eagle2-req-char}(a), \name{}
consistently outperforms SpS and EAGLE-2 by executing the draft model fewer
times and constructing a shallower tree, effectively adapting to the lower
acceptance length of EAGLE.
For example, as shown in \figref{appendix-llama-blackwell-eagle2-req-char}(b)
and (c), \name{} runs the draft model only 4.9 times for HumanEval (achieving an
acceptance length of 3.61), whereas it executes the draft model 7.1 times when
using EAGLE-3 as a draft model to achieve an acceptance length of 5.71.

\putssec{appendix-consumer-GPU}{Extension to Consumer-Grade GPU}
To demonstrate generalizability, we also provide further evaluations on a
consumer-grade GPU, the RTX 3090, similar to that done on RTX Pro 6000 Blackwell
in \ssecref{main-results}.
We evaluate a narrower range of request rates on the 3090 than on the RTX Pro
6000 Blackwell, because the smaller GPU memory capacity of the 3090 limits the
maximum allowable batch size.

\myparagraph{Performance Comparison with Varying Request Rates}

\begin{figure}[htbp]
    {
        \begin{center}
            \includegraphics[width=\textwidth]{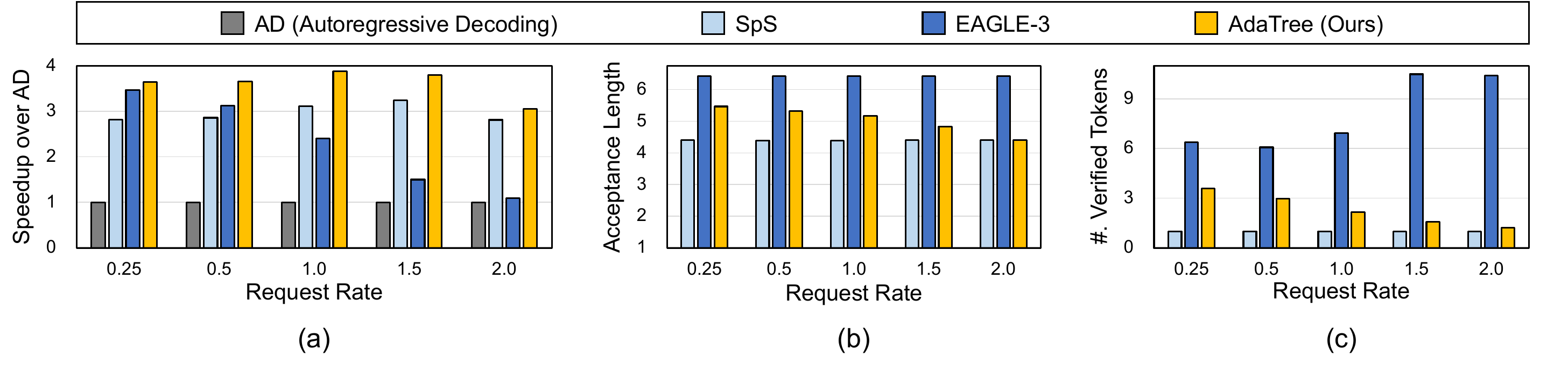}
        \end{center}
        \vspace{-15pt}
    }
    \caption{Performance comparison with varying request rates using GSM8K
      dataset on RTX 3090.
      (a) Speedup of SpS, EAGLE-3, and \name{} over AD.
      (b) Average acceptance length.
      (c) Average number of tokens processed per target model iteration,
      normalized to that of SpS.}
    \label{fig:appendix-llama-3090-eagle3-req-rate}
\end{figure}

As shown in \figref{appendix-llama-3090-eagle3-req-rate}, \name{} consistently
outperforms SpS and EAGLE-3 across all request rates by dynamically adjusting
the tree size.
A key difference on consumer-grade hardware is that \name{} constructs smaller
trees compared to the high-end GPU at the same request rate (e.g., size of 49 on
Pro 6000 Blackwell and 33 on 3090 at 0.5 requests/s).
This is because the ratio of compute capability (i.e., TFLOPS of the tensor
core) to the memory bandwidth is lower on the 3090 GPU, and thus, the saturation
to a compute-bound regime starts at smaller tree sizes.

\myparagraph{Performance Comparison with Varying Request Characteristics}

\begin{figure}[htbp]
    {
        \begin{center}
            \includegraphics[width=\textwidth]{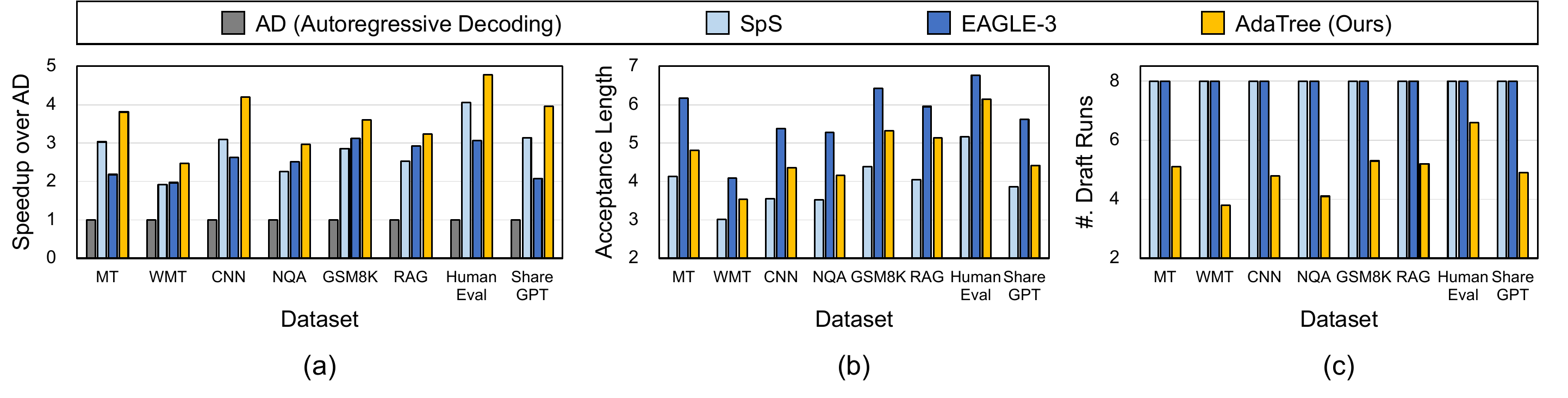}
        \end{center}
        \vspace{-15pt}
    }
    \caption{Performance comparison with varying datasets on RTX 3090 at request
      rate of 0.5.
      (a) Speedup of SpS, EAGLE-3, and \name{} over AD.
      (b) Average acceptance length.
      (c) Average number of draft model runs.}
    \label{fig:appendix-llama-3090-eagle3-req-char}
\end{figure}

As shown in \figref{appendix-llama-3090-eagle3-req-char}, \name{} consistently
outperforms SpS and EAGLE-3 across all datasets by dynamically adjusting the
tree depth.
Since the acceptance length, a metric tied to the request characteristics, is
independent of the underlying hardware, the performance trends remain consistent
with those observed for the Pro 6000 Blackwell in \sssecref{eval-request-char}
(\figref{eval-llama-blackwell-eagle3-req-char}).

\putssec{appendix-overhead}{Offline and Runtime Execution Overhead of \name{}}
We discuss the offline and runtime execution overhead of integrating \name{}'s
throughput estimation into an LLM serving system.

\myparagraph{Offline Overhead of \name{}}

For a given target model and hardware platform (e.g., GPU), we profile the model
execution time for varying batch sizes, context lengths, and input lengths,
which serve as training data for XGBoost.
Profiling takes approximately two hours per combination of target model and GPU
used in our evaluation.
As XGBoost is a lightweight machine learning model, training XGBoost only takes
a few seconds.
Note that this profiling is a one-time cost, and once XGBoost is trained, it is
used continuously in serving systems where batch sizes and request patterns keep
fluctuating.

\myparagraph{Runtime Execution Overhead of \name{}}

To demonstrate the execution overhead of our new throughput estimation module,
we present an execution time breakdown of a model iteration.
We evaluate three request rates of 0.5, 2.0, and 6.0 requests/s (average batch
sizes: 1.1, 2.3, and 8.7) using the GSM8K dataset on the Pro 6000 Blackwell GPU.

\begin{figure}[htbp]
    {
        \begin{center}
            \includegraphics[width=\textwidth]{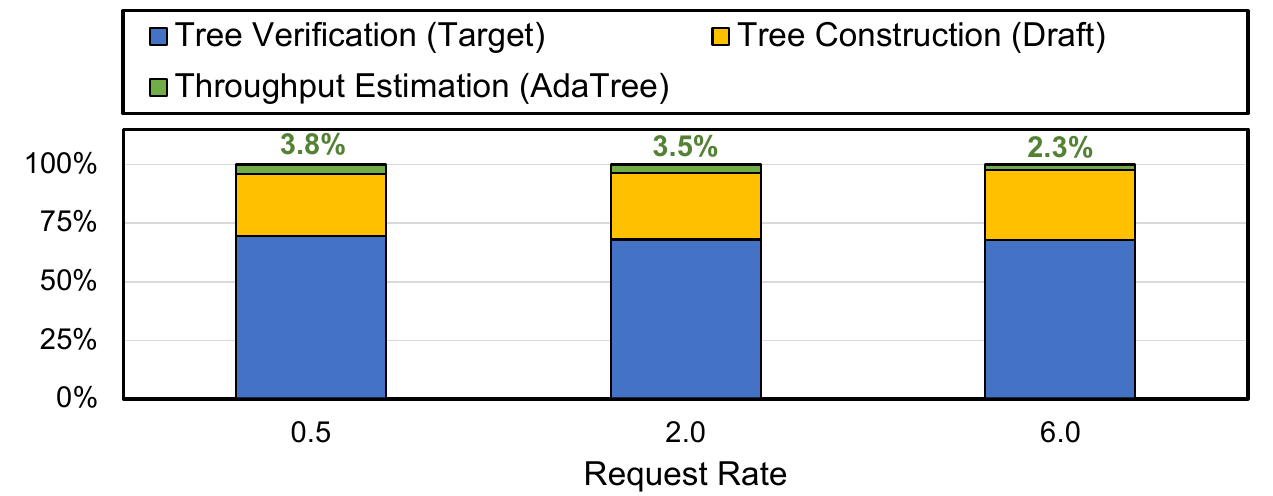}
        \end{center}
        \vspace{-15pt}
    }
    \caption{Latency breakdown of a single model iteration when \name{} is
      integrated.}
    \label{fig:appendix-latency-breakdown}
\end{figure}

Throughput estimation takes less than 4\% of the total execution time.
Notably, it constitutes a smaller ratio as the request rate and resulting batch
size increase, because the execution time of throughput estimation does not
scale with batch size, unlike model operations.
This highlights the scalability of \name{} as a lightweight throughput
estimation module for online serving, where continuous
batching~\cite{yu:jeo22, kwo:li23} is employed.

\putssec{appendix-adaptive-baseline}{Comparison to Other Adaptive Speculative Decoding}
We compare \name{} to two adaptive speculative decoding designs,
Tetris~\cite{wu:zho25} and OPT-Tree~\cite{wan:su25}.
Since Tetris adjusts chain-based speculative decoding by constraining the total
chain length to a fixed budget, we extend it to tree decoding for comparison,
denoting it as Tetris+Tree, by capping the total tree size at the GPU's
theoretical knee point.
For OPT-Tree, which stops expansion when the increase in cumulative probability
falls below a fixed threshold, we set the threshold to 0.8 as in the original
paper and select the best tree size by sweeping over values.
We report speedup over AD using the GSM8K dataset for varying request rates.

\begin{table}[t]
    \centering
    \caption{Speedup over AD for adaptive baselines.}
    \label{tab:appendix-adaptive-baseline}
    \small
    \begin{tabular}{lcccc}
        \toprule
        Request Rate & EAGLE-3 & \textbf{AdaTree} & Tetris+Tree & OPT-Tree \\
        \midrule
        2.0 reqs/s  & 4.00 & \textbf{4.09} & 3.62 & 3.44 \\
        3.0 reqs/s  & 4.19 & \textbf{4.66} & 4.25 & 4.00 \\
        4.0 reqs/s  & 4.71 & \textbf{5.98} & 5.47 & 5.02 \\
        \bottomrule
    \end{tabular}
\end{table}

As shown in \tabref{appendix-adaptive-baseline}, \name{} consistently
outperforms both adaptive baselines across varying serving conditions for the
following reasons.

First, for Tetris+Tree, setting the number of verified tokens to the theoretical
knee point is not always optimal.
The acceptance length does not grow linearly (\figref{char-system-load}(b)), so
pushing too many tokens beyond a certain point is wasteful.
In addition, the computation cost still slowly increases in the bandwidth-bound
region (\figref{design-model-exec-time-blackwell}(a)) because of a term
proportional to the total tree size (e.g., activation reads), and the growing
tree size also results in additional drafting cost.

Second, a static confidence threshold in OPT-Tree is particularly suboptimal in
batched serving.
When the batch size is small, relaxing the threshold is better because the
verification cost is relatively cheap in the bandwidth-bound region, making it
beneficial to push more tokens.
However, when the batch size increases, a stricter threshold is required because
the verification cost becomes more expensive.

In contrast, \name{} optimizes per-iteration throughput by jointly considering
the acceptance probability and execution time of target and draft models using
XGBoost.

\newpage
\section*{NeurIPS Paper Checklist}

\begin{enumerate}
\item {\bf Claims}
    \item[] Question: Do the main claims made in the abstract and introduction
      accurately reflect the paper's contributions and scope?
    \item[] Answer: \answerYes{}
    \item[] Justification: The claims in the abstract and introduction are
      discussed in detail throughout the main body and evaluation.
\item {\bf Limitations}
    \item[] Question: Does the paper discuss the limitations of the work
      performed by the authors?
    \item[] Answer: \answerYes{}
    \item[] Justification: We discuss the limitations of our work in conclusion
      (Section 6).
\item {\bf Theory assumptions and proofs}
    \item[] Question: For each theoretical result, does the paper provide the
      full set of assumptions and a complete (and correct) proof?
    \item[] Answer: \answerNA
    \item[] Justification: The paper does not include theoretical results.
    \item {\bf Experimental result reproducibility}
    \item[] Question: Does the paper fully disclose all the information needed
      to reproduce the main experimental results of the paper to the extent that
      it affects the main claims and/or conclusions of the paper (regardless of
      whether the code and data are provided or not)?
    \item[] Answer: \answerYes
    \item[] Justification: We tried to include all the necessary details, such
      as dataset, model types, and GPUs, to implement our proposal and reproduce
      the main results.
\item {\bf Open access to data and code}
    \item[] Question: Does the paper provide open access to the data and code,
      with sufficient instructions to faithfully reproduce the main experimental
      results, as described in supplemental material?
    \item[] Answer: \answerYes{}
    \item[] Justification: The code is publicly available at
      \url{https://github.com/snu-comparch/AdaTree}.
\item {\bf Experimental setting/details}
    \item[] Question: Does the paper specify all the training and test details
      (e.g., data splits, hyperparameters, how they were chosen, type of
      optimizer) necessary to understand the results?
    \item[] Answer: \answerYes
    \item[] Justification: We discuss those details in Section 5.1 Setting.
\item {\bf Experiment statistical significance}
    \item[] Question: Does the paper report error bars suitably and correctly
      defined or other appropriate information about the statistical
      significance of the experiments?
    \item[] Answer: \answerNo
    \item[] Justification: The experimental results reported in the paper are
      all model execution time, which we conducted in an isolated GPU system.
      Thus, we observe almost no variance in the measured times and did not
      report related mean or standard deviations.
\item {\bf Experiments compute resources}
    \item[] Question: For each experiment, does the paper provide sufficient
      information on the computer resources (type of compute workers, memory,
      time of execution) needed to reproduce the experiments?
    \item[] Answer: \answerYes
    \item[] Justification: The experiments are conducted in a single GPU system,
      which we discuss in Section 5.1.
\item {\bf Code of ethics}
    \item[] Question: Does the research conducted in the paper conform, in every
      respect, with the NeurIPS Code of Ethics
      \url{https://neurips.cc/public/EthicsGuidelines}?
    \item[] Answer: \answerYes
    \item[] Justification: The research conforms with the NeurIPS Code of
      Ethics.
\item {\bf Broader impacts}
    \item[] Question: Does the paper discuss both potential positive societal
      impacts and negative societal impacts of the work performed?
    \item[] Answer: \answerNA
    \item[] Justification: There is no societal impact of the work performed.
\item {\bf Safeguards}
    \item[] Question: Does the paper describe safeguards that have been put in
      place for responsible release of data or models that have a high risk for
      misuse (e.g., pre-trained language models, image generators, or scraped
      datasets)?
    \item[] Answer: \answerNA
    \item[] Justification: We do not release data or models.
\item {\bf Licenses for existing assets}
    \item[] Question: Are the creators or original owners of assets (e.g., code,
      data, models), used in the paper, properly credited and are the license
      and terms of use explicitly mentioned and properly respected?
    \item[] Answer: \answerYes
    \item[] Justification: We properly referenced the codes or methods from a
      third party in Section 5.1.
\item {\bf New assets}
    \item[] Question: Are new assets introduced in the paper well documented and
      is the documentation provided alongside the assets?
    \item[] Answer: \answerNA
    \item[] Justification: This work does not release new assets.
\item {\bf Crowdsourcing and research with human subjects}
    \item[] Question: For crowdsourcing experiments and research with human
      subjects, does the paper include the full text of instructions given to
      participants and screenshots, if applicable, as well as details about
      compensation (if any)?
    \item[] Answer: \answerNA
    \item[] Justification: The paper does not involve crowdsourcing nor research
      with human subjects.
\item {\bf Institutional review board (IRB) approvals or equivalent for research
  with human subjects}
    \item[] Question: Does the paper describe potential risks incurred by study
      participants, whether such risks were disclosed to the subjects, and
      whether Institutional Review Board (IRB) approvals (or an equivalent
      approval/review based on the requirements of your country or institution)
      were obtained?
    \item[] Answer: \answerNA
    \item[] Justification: This paper does not involve research with human
      subjects.
\item {\bf Declaration of LLM usage}
    \item[] Question: Does the paper describe the usage of LLMs if it is an
      important, original, or non-standard component of the core methods in this
      research?
      Note that if the LLM is used only for writing, editing, or formatting
      purposes and does \emph{not} impact the core methodology, scientific
      rigor, or originality of the research, declaration is not required.
    \item[] Answer: \answerNA
    \item[] Justification: The core method development in this research does not
      involve LLMs as one of the components.
      Our work is about accelerating the LLM serving through the flexible use of
      speculative decoding technique.
      Thus, LLM only serves as the workload of our study.
      We describe the used LLMs in Section 5.1.
\end{enumerate}

\end{document}